\documentclass[runningheads]{llncs}
\usepackage{graphicx}

\usepackage{tikz}
\usepackage{comment}
\usepackage{amsmath,amssymb} 
\usepackage{color}
\usepackage{booktabs}
\usepackage{hyperref}

\usepackage{pifont}
\newcommand{\cmark}{\ding{51}}%
\newcommand{\etal}{\textit{et al.}}

\usepackage{rotating}
\usepackage[caption=false]{subfig}
\usepackage{multirow}
\usepackage{adjustbox}
\usepackage{wrapfig}
\usepackage{capt-of}
\usepackage{varwidth}
\usepackage{booktabs}
\newsavebox\tmpbox

\usepackage[accsupp]{axessibility}  

\begin{document}
\pagestyle{headings}
\mainmatter

\title{Multi-modal Text Recognition Networks: Interactive Enhancements between Visual and Semantic Features} 

\titlerunning{MATRN}
%
\author{Byeonghu Na\inst{1}\index{Na, Byeonghu}\and
Yoonsik Kim\inst{2}\index{Kim, Yoonsik}\and
Sungrae Park\inst{3}\index{Park, Sungrae}$^*$}
\authorrunning{B. Na et al.}
%
\institute{Dept. of Industrial \& Systems Engineering, KAIST,  \email{wp03052@kaist.ac.kr} \and
Clova AI Research, NAVER Corp.,  \email{yoonsik.kim90@navercorp.com} \and Upstage AI Research, Upstage, 
\email{sungrae.park@upstage.ai}}
\maketitle

\begin{abstract}

Linguistic knowledge has brought great benefits to scene text recognition by providing semantics to refine character sequences. However, since linguistic knowledge has been applied individually on the output sequence, previous methods have not fully utilized the semantics to understand visual clues for text recognition. This paper introduces a novel method, called \textbf{M}ulti-mod\textbf{A}l \textbf{T}ext \textbf{R}ecognition \textbf{N}etwork (\textbf{MATRN}), that enables interactions between visual and semantic features for better recognition performances. Specifically, MATRN identifies visual and semantic feature pairs and encodes spatial information into semantic features. Based on the spatial encoding, visual and semantic features are enhanced by referring to related features in the other modality. Furthermore, MATRN stimulates combining semantic features into visual features by hiding visual clues related to the character in the training phase. Our experiments demonstrate that MATRN achieves state-of-the-art performances on seven benchmarks with large margins, while naive combinations of two modalities show less-effective improvements. Further ablative studies prove the effectiveness of our proposed components. Our implementation is available at \url{https://github.com/wp03052/MATRN}.
\end{abstract}

\section{Introduction}

Scene text recognition (STR), a major component of optical character recognition (OCR) technology, identifies a character sequence in a given text image patch (e.g. words in a traffic sign). Applications of deep neural networks have brought great improvements in the performance of STR models~\cite{Baek_2019_ICCV_CombBest,CRNN,ASTER,Wang_2020_DAN,Yu_2020_CVPR_SRN,RobustScanner}. They typically consist of a visual feature extractor, abstracting the image patch, and a character sequence generator, responsible for character decoding. Despite wide explorations to find better visual feature extractors and character sequence generators, existing methods still suffer from challenging environments: occlusion, blurs, distortions, and other artifacts~\cite{Baek_2019_ICCV_CombBest,JVSR}.

To address these remaining challenges, STR methods have tried to utilize linguistic knowledge on the output character sequence. The mainstream of the approaches had been to model recursive operations learning linguistic knowledge for next character prediction. RNN~\cite{Baek_2019_ICCV_CombBest,ASTER} and Transformer~\cite{SATRN,NRTR,Holistic} have been applied to learn the auto-regressive language model (LM). 
However, the auto-regressive process requires expensive repeated computations and also learns limited linguistic knowledge from the uni-directional transmission. 

To compensate for the issues, Yu \etal~\cite{Yu_2020_CVPR_SRN} propose SRN that refines an output sequence without auto-regressive operations. After identifying a seed character sequence, SRN re-estimates the character for each position at once by utilizing a Transformer encoder with a subsequent mask.
Based on SRN, Fang \etal~\cite{ABINet} improve the iterative refinement stages by explicitly dividing a vision model (VM) and an LM by blocking gradient flows and employing a bi-directional LM pre-trained on unlabeled text datasets. These methods incorporating semantic knowledge of LMs provide breakthroughs in recognizing challenging examples with ambiguous visual clues. 
However, the character refinements without visual features might lead to wrong answers by missing existent visual clues.

For better combinations of semantics and visual clues, Bhunia \etal~\cite{JVSR} propose a multi-stage decoder referring to visual features multiple times to enhance semantic features. At each stage, a character sequence, designed as differentiable with Gumbel-softmax, is re-fined by re-assessing visual clues.
Concurrently, Wang \etal~\cite{VisionLAN} propose VisionLAN utilizing a language-aware visual mask that occludes selected character regions for enhancing the visual clues at the training phase.
They prove that combining visual clues and semantic knowledge leads to better STR performances. 
Inspired by them, we raise a novel question: what is the best way to model the interactions between visual and semantic features identified by VM and LM, respectively?

\begin{figure}[t]
    \centering
    \subfloat[]
    {
        \includegraphics[width=0.23\linewidth]{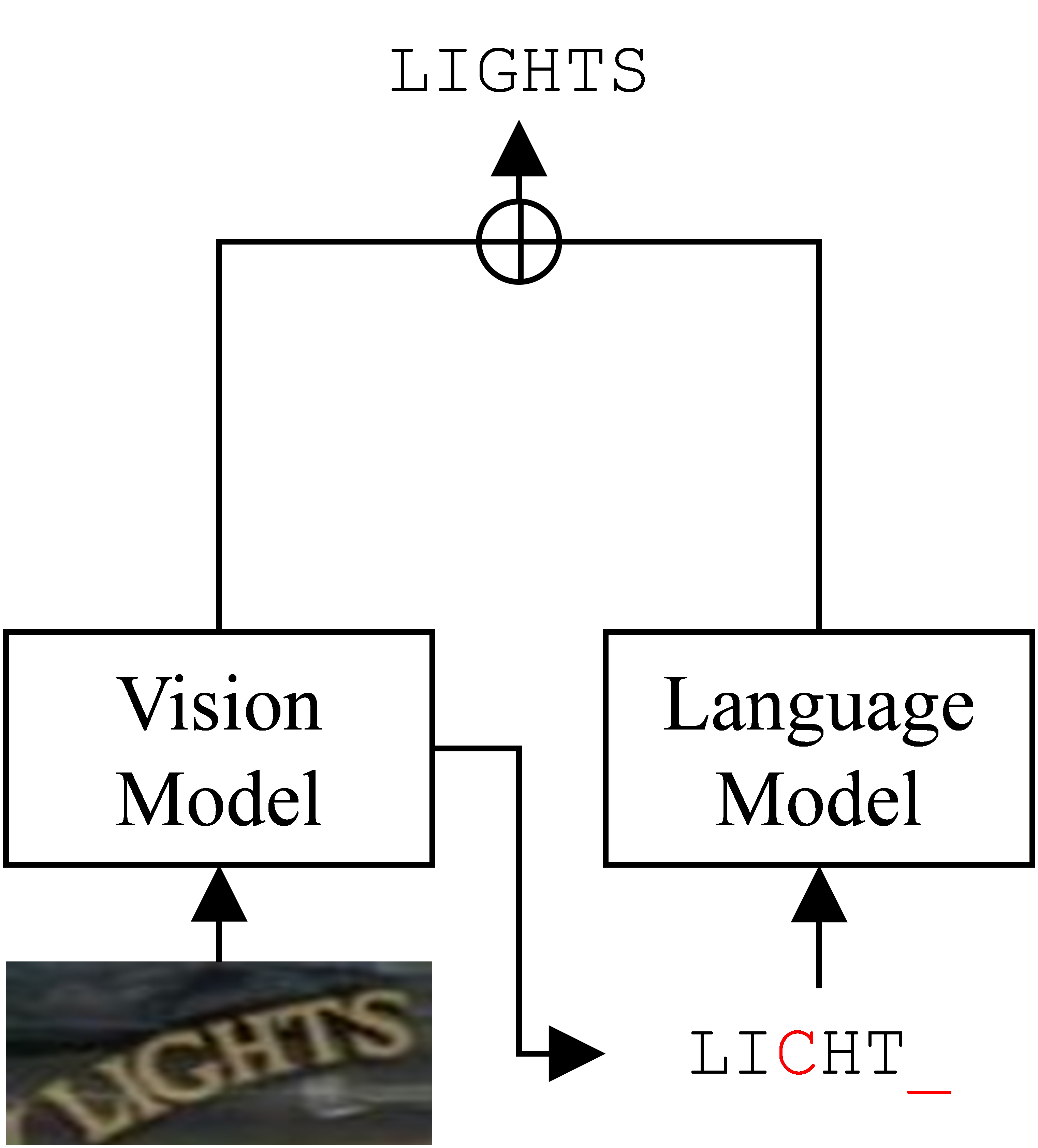}
        \label{fig:SRN}
    }
    \subfloat[][]
    {
        \includegraphics[width=0.23\linewidth]{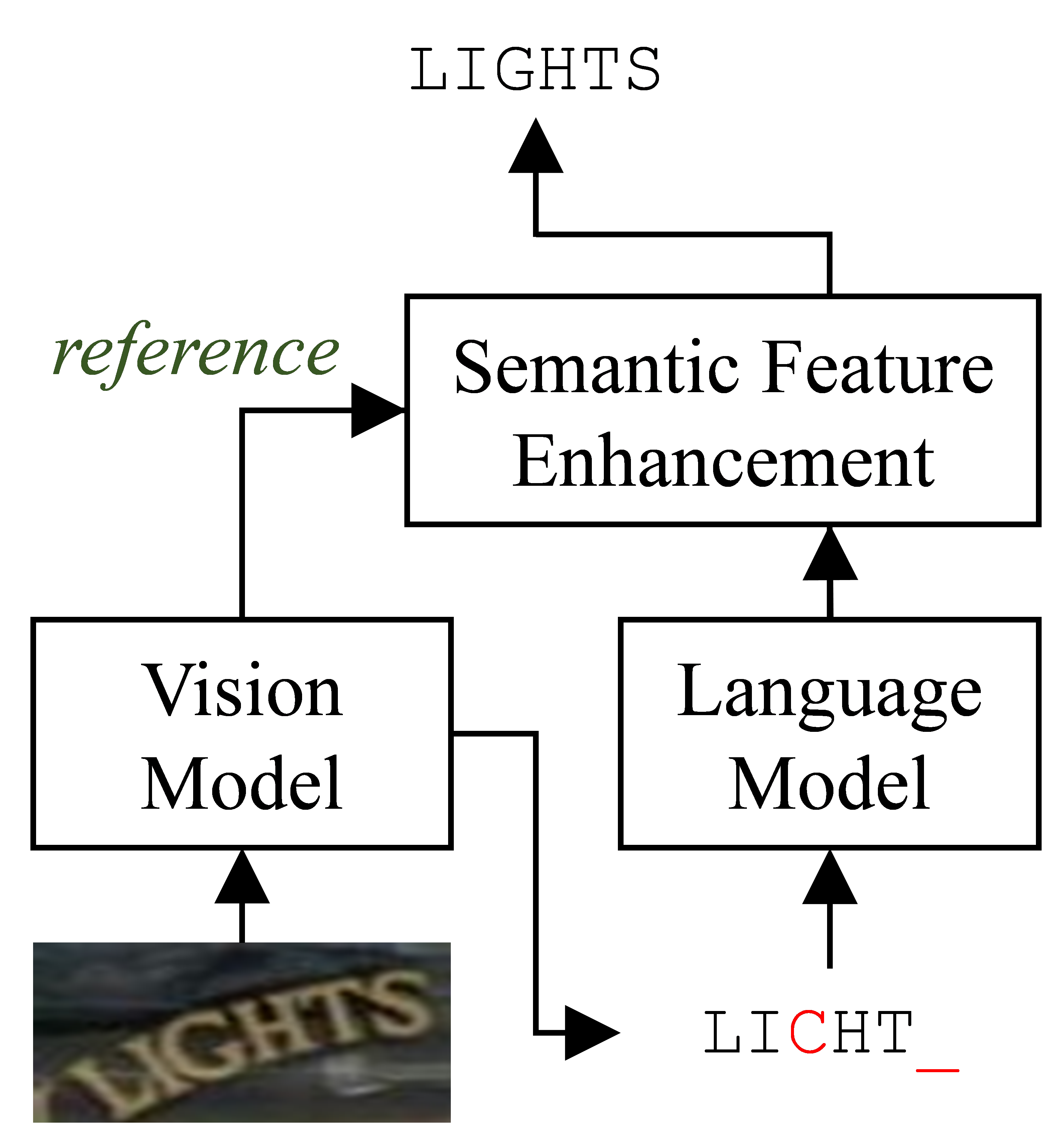}
        \label{fig:V2S}
    }
    \subfloat[]
    {
        \includegraphics[width=0.23\linewidth]{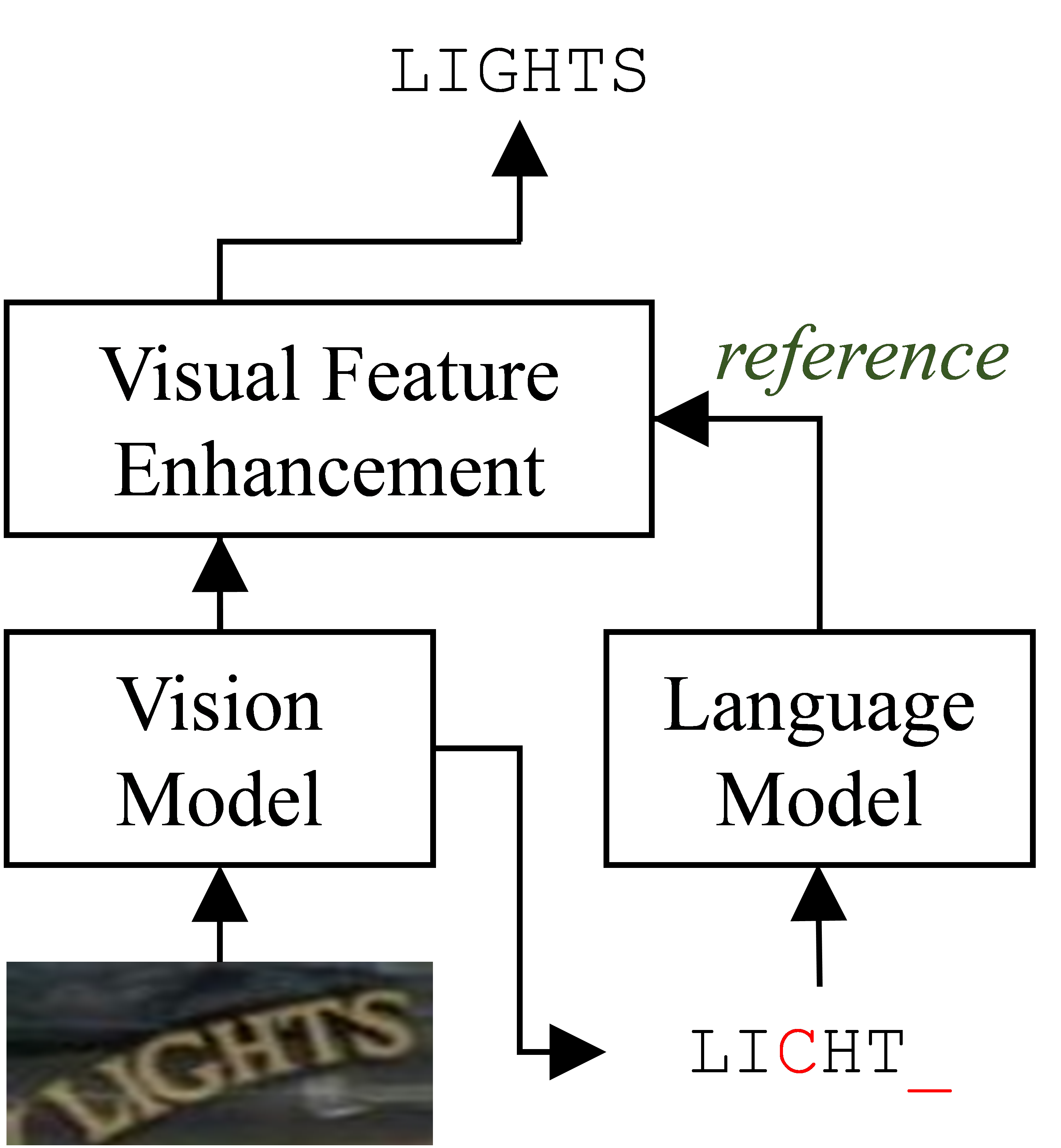}
        \label{fig:S2V}
    }
    \subfloat[]
    {
        \includegraphics[width=0.23\linewidth]{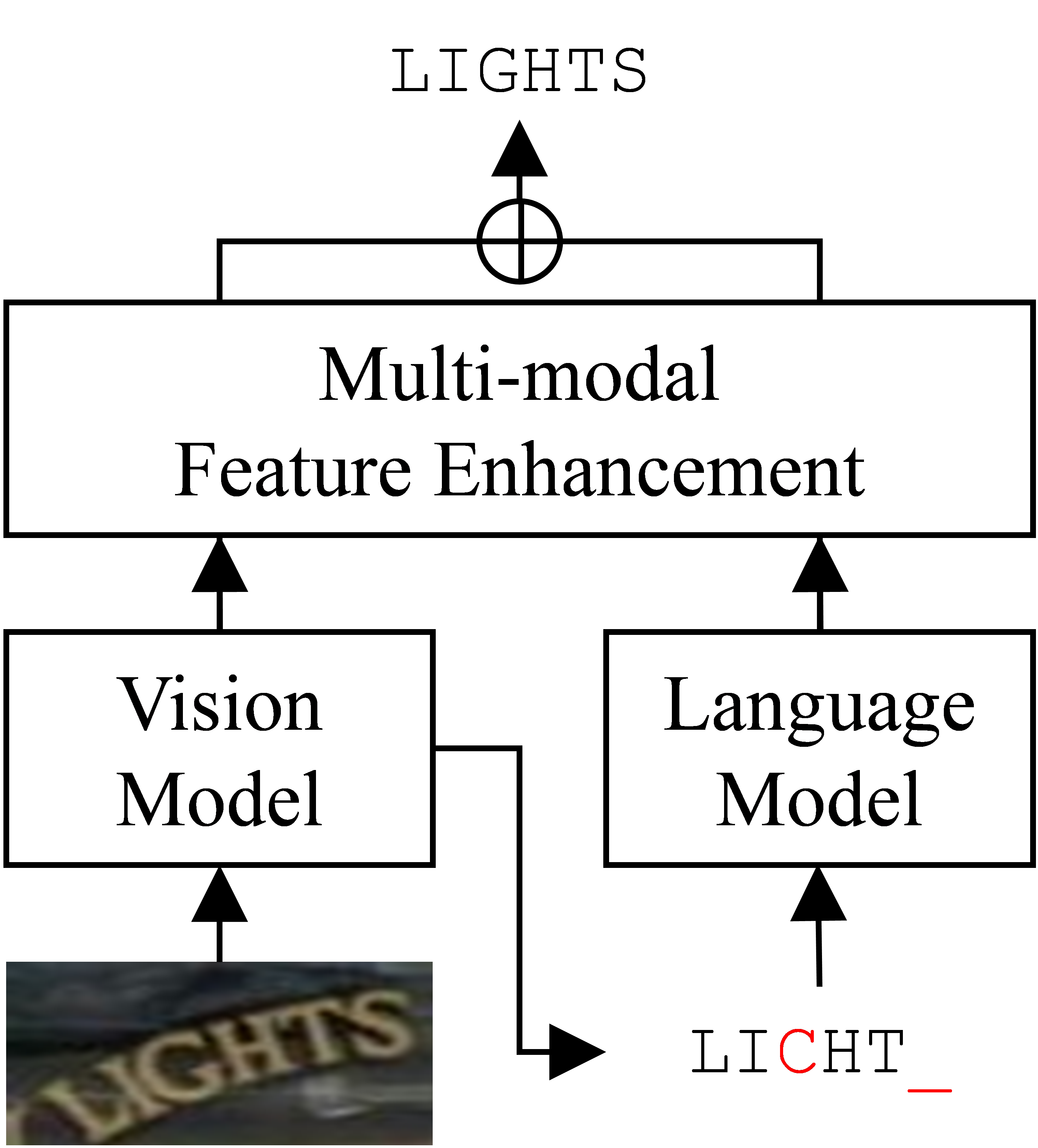}
        \label{fig:MATRN}
    }
    \label{fig:Models}
    \caption{
    Four types of visual and semantic feature fusions for STR: (a) simple combination of outputs from VM and LM, (b) visual-to-semantic feature fusion, (c) semantic-to-visual feature fusion, and (d) bi-directional feature fusion. SRN~\cite{Yu_2020_CVPR_SRN} is placed in (a) by applying LM to refine the output of VM. ABINet~\cite{ABINet}, PIMNet~\cite{PIMNet}, and JVSR~\cite{JVSR} can be aligned in (b) because their decoders refer to visual features iteratively during refining the final output sequence. VisionLAN~\cite{VisionLAN} combines semantic information into visual features in a similar way to (c). Our method, MATRN, is positioned in (d) by enhancing both features through the bi-directional reference.}
\end{figure}

To answer the question, this paper introduces a simple-but-effective extension of a STR model, named \textbf{M}ulti-mod\textbf{A}l \textbf{T}ext \textbf{R}ecognition \textbf{N}etwork (\textbf{MATRN}), that enhances visual and semantic features by referring to features in both modalities. MATRN consists of three proposed modules applied upon visual and semantic features: (1) multi-modal feature enhancement, incorporating bi-modalities to enhance each feature, (2) spatial encoding for semantics, linking two different modalities, (3) visual clue masking strategy, stimulating the cross-references between visual and semantic features. Figure~\ref{fig:Models} shows four types of visual and semantic feature fusions. MATRN is positioned in the bi-directional feature fusion (Figure~\ref{fig:MATRN}) by applying multi-modal feature enhancement. To the best of our knowledge, this natural but simple extension has never been explored.  

The resulting model, MATRN, is architecturally simple but effective. In addition, the visual and semantic feature fusions are not expensive because the whole process is conducted in parallel. When we evaluate simple combinations of visual and semantic features without our proposed components, the performance improvements are observed as less-effective. However, interestingly, the proposed components contribute to STR performances effectively and lead MATRNs to achieve superior performances with notable gains from the current state-of-the-art. Consequently, our paper proves that semantics is helpful to capture better visual clues as well as that combining visual and semantic features reaches better STR performances.

Our contributions are threefold.
\begin{itemize}
    \item We explore the combinations of visual and semantic features, identified by VM and LM, and prove their benefits. To the best of our knowledge, multi-modal feature enhancements with bi-directional fusions are novel components, that are natural extensions but have never been explored. 
    \item We propose a STR method, named MATRN, that contains three major components, spatial encoding for semantics, multi-modal feature enhancements, and visual clue masking strategy, for better combinations of two modalities. Thanks to the effective contributions of the proposed components, MATRN achieves state-of-the-art performances on seven STR benchmarks. 
    \item We provide empirical analyses that illustrate how our components improve STR performances as well as how MATRN contributes to the existing challenges.
\end{itemize}

\section{Related Work}

To utilize the benefits of a bi-directional Transformer, the non-autoregressive decoder has been introduced in the STR community. The general decoding process of them~\cite{ViTSTR,ABINet,PIMNet,VisionLAN,Yu_2020_CVPR_SRN} lies in the effective construction of a sequence processed parallelly in the decoder. Specifically, positional embeddings describing the order of the output sequence are used to align visual (or semantic) features. Although the output sequence is generated in parallel, the bi-directional Transformer has shown comparable performances with the auto-regressive approaches.

ViTSTR~\cite{ViTSTR} mainly focused on their VM without explicitly learning the LM. Inspired by the success of ViT~\cite{ViT}, ViTSTR~\cite{ViTSTR} has adopted ViT training scheme to STR. Specifically, its composition is very simple composed of the Transformer encoder and is trained with un-overlapped patches. 

In order to incorporate linguistic knowledge, PIMNet~\cite{PIMNet}, SRN~\cite{Yu_2020_CVPR_SRN} and ABINet~\cite{ABINet} have been proposed. To learn linguistic knowledge from the auto-regressive model, PIMNet~\cite{PIMNet} proposed step-wise predictions and similarity distance between non-autoregressive and auto-regressive models. SRN and ABINet introduced a language modality that refines the output sequence of VM. Then, the final predictions are achieved by fusing the output sequences of LMs and VMs. In SRN~\cite{Yu_2020_CVPR_SRN}, the LM is trained along with VM where the LM learns semantic information from other words. Based on SRN, ABINet~\cite{ABINet} improves the iterative refinement stages by explicitly dividing the VM and LM. With a pretraining LM on unlabeled text datasets, it provides breakthroughs in recognizing challenging examples with ambiguous visual clues.

To interactively combine LM and VM, multi-modal recognizers are also introduced~\cite{JVSR,VisionLAN}.
JVSR~\cite{JVSR} proposes a multi-stage decoder referring to visual features multiple times to enhance semantic features. 
Specifically, it is based on an RNN-attention decoder with multi-stages where each stage generates an output sequence and visual features are employed for updating each hidden state. 
Since the decoder takes a hidden state as an input, the visual feature can iteratively enhance the semantic features.
Concurrently, VisionLAN~\cite{VisionLAN} proposes a language-aware visual mask that refers to semantic features for enhancing the visual features. Given a masking position of the word, the masking module occludes corresponding visual feature maps of the character region at the training phase.
The previous multi-modal recognizers focus on one modality for final prediction and they utilize the other modality to improve their chosen one. In contrast, we explore the multiple combinations of multi-modal processes and propose MATRN which conducts both bi-directional enhancements.

\section{MATRN}

\begin{figure*}[t!]
\centering
\includegraphics[width=0.99\linewidth]{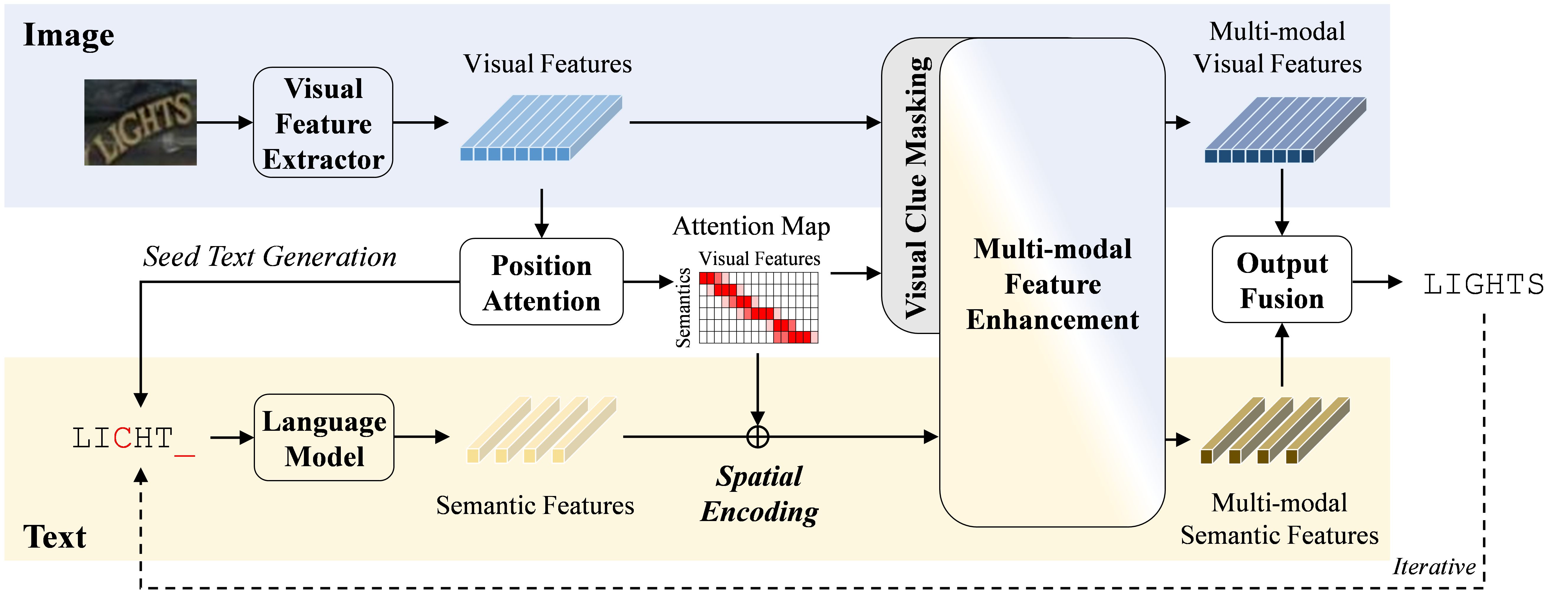}
\caption{An overview of MATRN. A visual feature extractor and an LM extract visual and semantic features, respectively. By utilizing the attention map, representing relations between visual features and character positions, MATRNs encode spatial information into the semantic features and hide visual features related to a randomly selected character. Through the multi-modal feature enhancement module, visual and semantic features interact with each other and the enhanced features in the two modalities are fused to finalize the output sequence. }
\label{fig:overview}
\end{figure*}

Here, we describe our recognition model, \emph{MATRN}, which incorporates both visual and semantic features. We will provide an overview of our method, and then describe each component in detail. 

\subsection{Overview of MATRNs}
Figure \ref{fig:overview} shows the overview of our model. It includes a visual feature extractor and a seed text generator to embed an image and provide an initial sequence of characters, as traditional STR models do. LM is applied to the seed text to extract semantic features.

Our contributions are focused on incorporating the visual and semantic features for better STR performances. Our method first encodes spatial positions into semantic features by utilizing an attention map identified during the seed text generation.
The multi-modal feature enhancement module enriches individual visual and semantic features by incorporating multi-modalities. The enhanced features are named as \textit{multi-modal visual features}, enhanced visual features with semantic knowledge, and \textit{multi-modal semantic features}, enhanced semantic features with visual clues. Finally, both features are combined to provide the output sequence.  

In the training phase of MATRN, the visual clue masking module hides visual features, related to a single character to stimulate the combination of semantics. Furthermore, the output sequence can be iteratively applied into the seed text for the LM, as follows \cite{ABINet}. 

\subsection{Visual and Semantic Feature Extraction}

To identify visual and semantic features, we construct three components: visual feature extractor, seed text generator, and language model. The following describes each module.

For the visual feature extractor, ResNet and Transformer units~\cite{ABINet,Yu_2020_CVPR_SRN} are applied. The ResNet, $F^{\text{V.R}}$, with 45 layers embeds an input image, $\mathbf{X} \in \mathbb{R}^{H \times W \times 3}$, into convolution visual features, $\mathbf{V}^{\text{conv}} \in \mathbb{R}^{\frac{H}{4} \times \frac{W}{4} \times D}$. $H, W$ are height and width of the image and $D$ is the feature dimension. Before applying the Transformer, sinusoidal spatial position embeddings, $\mathbf{P}^{\text{V}}\in \mathbb{R}^{\frac{H}{4} \times \frac{W}{4} \times D}$, are added. Then, the Transformer, $F^{\text{V.T}}$, with three layers is applied:
\begin{equation}
  \mathbf{V} =  F^{\text{V.T}}(F^{\text{V.R}}(\mathbf{X}) + \mathbf{P}^{\text{V}}),  
\end{equation}
where $\mathbf{V} \in \mathbb{R}^{\frac{H}{4} \times \frac{W}{4} \times D}$ indicates visual features that are the outputs of the visual feature extractor.

For the seed text generation, an attention mechanism is utilized to transcribe visual features into character sequences. Specifically, an attention map, $\mathbf{A}^{\text{V-S}} \in \mathbb{R}^{T \times \frac{HW}{16}}$, is calculated by setting queries as text positional embeddings, $\mathbf{P}^{\text{S}}\in \mathbb{R}^{T \times D}$, and keys as $\mathcal{G}(\mathbf{V})\in\mathbb{R}^{\frac{HW}{16} \times D}$ in the attention mechanism, where $T$ is the maximum length of sequence and $\mathcal{G}(\cdot)$ is a mini U-Net. Through the attention map, visual features are abstracted upon sequential features, $\mathbf{E}^{\mathbf{V}} =  \mathbf{A}^{\text{V-S}} \widetilde{\mathbf{V}}$, where $\widetilde{\mathbf{V}} \in\mathbb{R}^{\frac{HW}{16} \times D}$ indicates the flattened visual features. By applying a linear layer and softmax function, a seed character sequence, $\mathbf{Y}_{\text{(0)}}\in\mathbb{R}^{T \times C}$, is generated, where $C$ indicates the number of character classes. The whole process can be formalized as follows;
\begin{gather}
 \mathbf{A}^{\text{V-S}} =  \text{softmax}\left(\mathbf{P}^\text{S}\mathcal{G}(\mathbf{V})^{\top}/\sqrt{D}\right), \label{eq:Att} \\
\mathbf{Y}_{\text{(0)}} = \text{softmax}\left(\mathbf{A}^{\text{V-S}}\widetilde{\mathbf{V}}\mathbf{W}\right), \label{eq:V_Y}
\end{gather}
where $\mathbf{W}\in\mathbb{R}^{D \times C}$ indicates a linear transition matrix.

The LM, introduced by \cite{ABINet}, consists of four Transformer decoder blocks. 
It uses $\mathbf{P}^{\text{S}}$ as inputs and $\mathbf{Y}_{\text{(0)}}$ as the key/value of the attention layer. By processing the multiple decoder blocks, the LM identifies semantic features, $\mathbf{S} \in \mathbb{R}^{T \times D}$;
\begin{equation}
    \mathbf{S} = F^{\text{LM}}(\mathbf{Y}_{\text{(0)}}),
\end{equation}
where $F^{\text{LM}}$ indicates the LM. We initialize the LM with the weights, pre-trained on WikiText-103, provided by \cite{ABINet}.

\subsection{Spatial Encoding to Semantic Features}

One important point of combining the visual and semantic features is how to align each piece of information of different modalities. To guide the relationship between visual and semantic features, MATRN encodes spatial positions of visual features into semantic features. We call this process \emph{spatial encoding to semantics (SES)}. 

The key idea of SES is utilizing the attention map $\mathbf{A}^{\text{V-S}}$, used for the seed text generation, and the spatial position embedding $\mathbf{P}^\text{V}$, introduced in the visual feature extractor. Since $\mathbf{A}^{\text{V-S}}$ provides which visual features are used to estimate a character at each position, the spatial positions for semantic features, $\mathbf{P}^\text{Align} \in \mathbb{R}^{T \times D}$, are calculated as follows;
\begin{equation}
   \mathbf{P}^\text{Align} =  \mathbf{A}^{\text{V-S}}  \widetilde{\mathbf{P}}^{\text{V}},
\end{equation}
where $\widetilde{\mathbf{P}}^{\text{V}} \in \mathbb{R}^{\frac{HW}{16} \times D}$ is the flattened sinusoidal spatial position embedding, $\mathbf{P}^{\text{V}}$. Then, we encode the spatial information into semantic features:
\begin{equation}
   \mathbf{S^\text{Align}} = \mathbf{S} + \mathbf{P}^\text{Align}.
\end{equation}
From this encoding process, the spatially aligned semantic features, $\mathbf{S^\text{Align}}$, contain spatial clues of visual features are highly related. 
It should be noted that SES does not need any additional parameters as well as it is simple and effective for the cross-references between visual and semantic features.

\subsection{Multi-modal Features Enhancement}

Now, we hold visual features, $\tilde{\mathbf{V}}$, that learn visual clues for character estimations, and semantic features, $\mathbf{S^\text{Align}}$, that contain linguistic knowledge for a character sequence. Previous methods~\cite{Yu_2020_CVPR_SRN,ABINet} simply use a gated mechanism to seed character feature $\mathbf{E}^{\text{V}}$ and semantic feature $\mathbf{S}$. However, this simple fusion mechanism might not completely utilize these two features. Therefore, we propose a way in which the visual and semantic features refer to each other effectively and consequently enhance the features.

Multi-modal transformer~\cite{tsai2019MULT_MT}, which consists of transformer layers processing multiple types of features at once, has been introduced in several domains such as visual question answering~\cite{hu2020iterative_M4C}, vision-language navigation~\cite{chen2021hamt_VLN}, autonomous driving~\cite{prakash2021multi}, video retrieval~\cite{gabeur2020multi_VR}, and many others. Inspired by them, we employ the multi-modal transformer for visual and semantic features enhancement for STR. The multi-modal transformers have multiple Transformer encoder blocks that consist of an attention layer and a feed-forward layer. At the attention layer, both visual and semantic features are processed through self-attentions. Since the queries determine their major modality, visual features are enhanced as multi-modal visual features, $\mathbf{V}^\text{M} \in \mathbb{R}^{\frac{HW}{16} \times D}$, and semantic features are updated into multi-modal semantic features, $\mathbf{S}^\text{M} \in \mathbb{R}^{T \times D}$.

\subsection{Final Output Fusion}

Both multi-modal features are utilized to finalize the output character sequence. While multi-modal semantic features are already aligned as a sequence, multi-modal visual features are required to be re-organized to estimate characters. To align the visual features into a sequence, we apply a character generator, which has the same architecture of the seed text generator, to aggregate $\mathbf{V}^{\text{M}}$ into sequential features, $\mathbf{E}^{\mathbf{V}^{\text{M}}}$ (See \S3.2). Afterward, two sequential features, $\mathbf{E}^{\mathbf{V}^{\text{M}}}$ and $\mathbf{S}^\text{M}$, are combined through a gate mechanism to identify features, $\mathbf{F}\in\mathbb{R}^{T \times D}$, used for final character estimations:
\begin{align}
   \mathbf{G} & = \sigma \left ( \left [ \mathbf{E}^{\mathbf{V}^{\text{M}}}; \mathbf{S}^\text{M} \right ] \mathbf{W}^\text{gated} \right ), \\
   \mathbf{F} & = \mathbf{G} \odot \mathbf{E}^{\mathbf{V}^{\text{M}}} + (1-\mathbf{G}) \odot \mathbf{S}^\text{M},
\end{align}
where $\mathbf{W}^\text{gated} \in \mathbb{R}^{2D \times D}$ is a weight, $\left[;\right]$ indicates concatenation, and $\odot$ is element-wise product. Finally, a linear layer and softmax function are applied on $\mathbf{F}$ to estimate a character sequence, $\mathbf{Y}_\text{(1)}\in\mathbb{R}^{T \times C}$.

\begin{figure}[t]
    \centering
    \includegraphics[width=0.75\linewidth]{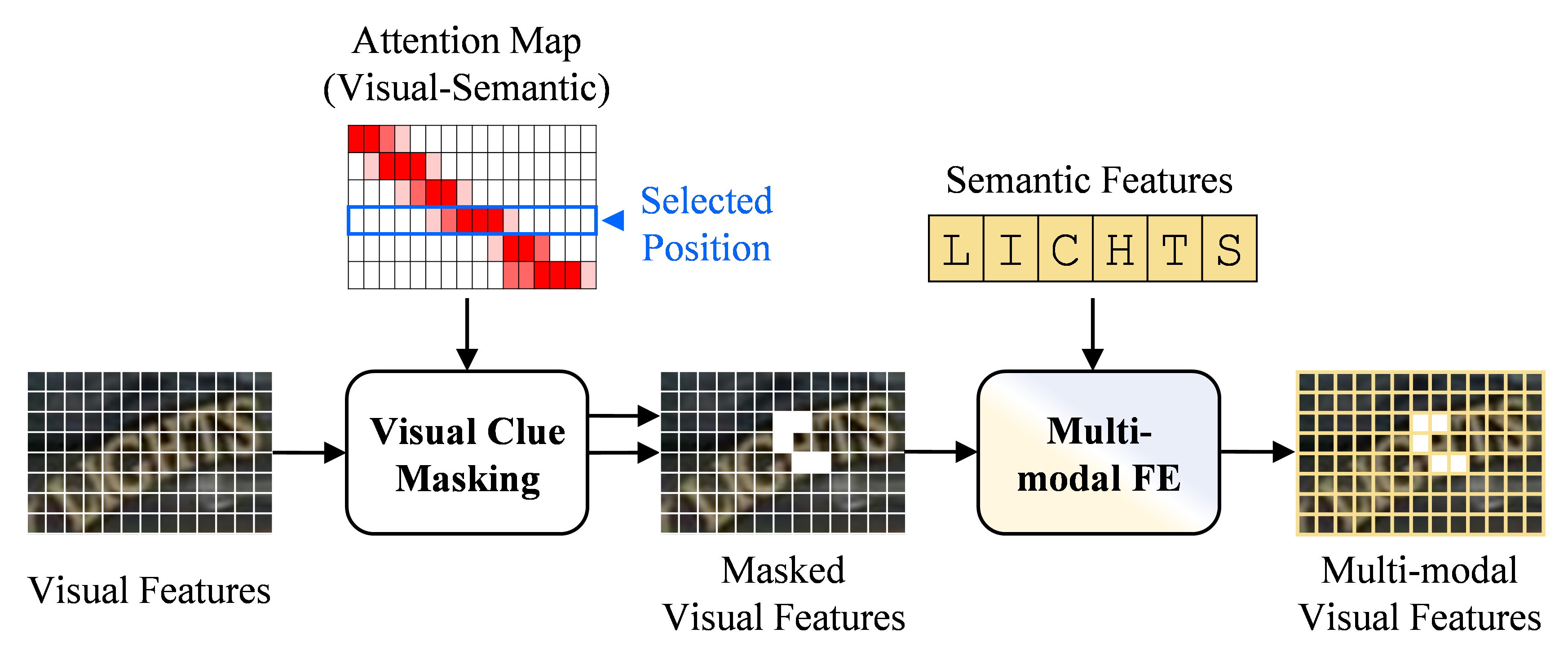}
    \caption{Conceptual descriptions of visual clue masking strategy. Based on the attention map, representing relations between visual features and characters, influential features for a randomly selected character position are masked. In the multi-modal FE stage, semantic features are stimulated to be merged more strongly to compensate for the missing visual clues.}
    \label{fig:Masking}
\end{figure}

\subsection{Visual Clue Masking Strategy}

To facilitate a better blend of the visual and semantic features, we propose a visual clue masking strategy, which is motivated by VisionLAN~\cite{VisionLAN}. This strategy selects a single character randomly and hides corresponding visual features based on the attention map, $\mathbf{A}^\text{V-S}$, identified in the seed text generation.
By explicitly deleting influential features for the character estimation, the multi-modal FE module becomes stimulated to encode semantic knowledge into the visual features in order to compensate for the missing visual clues. Figure~\ref{fig:Masking} provides the conceptual description of the visual clue masking strategy.

The masking process chooses a position randomly in a character sequence and finds the top-$K$ visual features relevant to the chosen position. For example, if the fourth position is selected, the process identifies the front $K$ visual features in the descending order of the attention scores at the fourth position. The identified visual features are replaced into $\mathbf{v}_{\text{[MASK]}}\in\mathbb{R}^{D}$. The visual clue masking is only applied in the training phase. To reduce the discrepancy between training and evaluating phases, we keep the identified features unchanged with probability 0.1, as like \cite{devlin-etal-2019-bert}.

\subsection{Training Objective}

MATRN is trained as end-to-end learning with multi-task cross-entropy objectives from multi-level visual and semantic features. We denote $\mathcal{L}_\mathbf{*}$ is a cross-entropy loss for estimated character sequences from a feature $*$. For the estimations, a linear layer and a softmax function are utilized.
In addition, MATRN applies iterative semantic feature correction to resolve the noisy input for the LM, as follows ~\cite{ABINet,lee_etal_2018_deterministic}. At the iteration, the input of LM is replaced into the output of the output fusion layer (See Figure~\ref{fig:overview}). The loss of MATRN is formed as follows;
\begin{align}
   \mathcal{L} & = \mathcal{L}_{\mathbf{E}^{\mathbf{V}}} + \frac{1}{M} \sum_{i=1}^{M} \left ( \mathcal{L}_{\mathbf{S}_{(i)}} + \mathcal{L}_{\mathbf{S}^\text{M}_{(i)}} + \mathcal{L}_{\mathbf{E}^{\mathbf{V}^\text{M}}_{(i)}} + \mathcal{L}_{\mathbf{F}_{(i)}} \right ),
\end{align}
where $M$ is the number of iterations. Here, ${\mathbf{S}_{(i)}}$, ${\mathbf{S}^\text{M}_{(i)}}$, ${\mathbf{E}^{\mathbf{V^\text{M}}}_{(i)}}$, and ${\mathbf{F}_{(i)}}$ indicate the semantic, multi-modal semantic, multi-modal visual, and final fused features at the $i$-th iteration, respectively.   

\section{Experiments}

\subsection{Datasets}
For fair comparisons, we use the same training dataset and evaluation protocol with~\cite{ABINet,Yu_2020_CVPR_SRN}. For the training set, we use two widely-used synthetic datasets, MJSynth~\cite{Jaderberg14c_MJSynth} and SynthText~\cite{Gupta16_SynthText}. MJSynth has 9M synthetic text images and SynthText consists of 7M images including examples with special characters. Most previous works have used these two synthetic datasets together: MJSynth + SynthText~\cite{Baek_2019_ICCV_CombBest}.

For evaluation, eight widely used real-world STR benchmark datasets are used as test datasets. The datasets are categorized into two groups: ``regular'' and ``irregular'' datasets, according to the geometric layout of texts.  ``regular'' datasets mainly contain horizontally aligned text images. IIIT5K (IIIT)~\cite{IIIT5k} consists of 3,000 images collected from the web. Street View Text (SVT)~\cite{SVT} has 647 images collected from Google Street View. ICDAR2013 (IC13)~\cite{IC13} represents images cropped from mall pictures and has two variants; 857 images (IC13$_\text{S}$) and 1015 images (IC13$_\text{L}$). We utilize all two variants for providing fair comparisons. We skipped the evaluation  on ICDAR2003~\cite{IC03} because it contains duplicated images with IC13~\cite{Baek_2019_ICCV_CombBest}.

``irregular'' datasets consist of more examples of text in arbitrary shapes. 
ICDAR2015 (IC15) consists of images taken from scenes and also has two versions; 1,811 images (IC15$_\text{S}$) and 2,077 images (IC15$_\text{L}$). Street View Text Perspective (SVTP)~\cite{SVTP} contains 645 images of which texts are captured in perspective views. CUTE80 (CUTE)~\cite{CUTE80} consists of 288 images of which texts are heavily curved. 

In our analyses, we measure the word prediction accuracy on each dataset.
For ``Total.'', we evaluate the accuracy of unified evaluation datasets except for IC13$_\text{L}$ and IC15$_\text{L}$. It should be noted that we follow the philosophy of Baek \etal~\cite{Baek_2019_ICCV_CombBest} which compares STR models upon the common evaluation datasets.

\subsection{Implementation Details}
\label{sec:implementaion}

The height and width of the input image are 32 and 128 by re-sizing text images and we apply image augmentation methods \cite{ABINet,lyu2018mask,Yu_2020_CVPR_SRN} such as rotation, color jittering, and noise injection. The number of character classes is 37; 10 for digits, 26 for alphabets, and a single padding token.

We borrow the network structures of the visual feature extractor, seed text generator, and language model from ABINet~\cite{ABINet}. We set the feature dimension, $D$, as 512 and the maximum length of the sequence, $T$, as 25. For the multi-modal transformer, we use 2 Transformer blocks with 8 heads and a hidden size of 512. The iteration number $M$ is set to 3 unless otherwise specified. We fixed the number of visual features mask, $K$, as 10.

We adopt the code from ABINet\footnote{https://github.com/FangShancheng/ABINet}, and keep the experiment configuration. We use a pre-trained visual feature extractor and a pre-trained language model, which are provided by~\cite{ABINet}. We use 4 NVIDIA GeForce RTX 3090 GPUs to train our models with batch size of 384. We used Adam~\cite{Adam} optimizer of initial learning rate $10^{-4}$, and the learning rate is decayed to  $10^{-5}$ after six epochs.

\subsection{Comparison to State-of-the-Arts}

\begin{table}[t]
\caption{Recognition accuracies (\%) on eight benchmark datasets, including the variant versions. The underlined values represent the best performances among the previous STR methods and the bold values indicate the best performances among all models including ours. For our implementation, we conduct repeated experiments with three different random seeds and report the averaged accuracy with standard deviation.}
\label{tab:accuracy}
\small
\tabcolsep=0.1cm
\centering
\begin{centering}
\begin{tabular}{ll|cccc|cccc}
\toprule
 & & \multicolumn{4}{c|}{\textbf{Regular test dataset}} & \multicolumn{4}{c}{\textbf{Irregular test dataset}} \\
 \textbf{Model} & \textbf{Year} & IIIT & SVT & IC13$_\text{S}$ & IC13$_\text{L}$ & IC15$_\text{S}$ & IC15$_\text{L}$ & SVTP & CUTE \\
\midrule
 CombBest~\cite{Baek_2019_ICCV_CombBest} & 2019 & 87.9 & 87.5 & 93.6 & 92.3 & 77.6 & 71.8 & 79.2 & 74.0 \\
 ESIR~\cite{Zhan_2019_CVPR_ESIR} & 2019 & 93.3 & 90.2 & - & 91.3 & - & 76.9 & 79.6 & 83.3 \\
 SE-ASTER~\cite{Qiao_2020_CVPR} & 2020 & 93.8 & 89.6 & - & 92.8 & 80.0 & & 81.4 & 83.6 \\
 DAN~\cite{Wang_2020_DAN} & 2020 & 94.3 & 89.2 & - & 93.9 & - & 74.5 & 80.0 & 84.4 \\
 RobustScanner~\cite{RobustScanner} & 2020 & 95.3 & 88.1 & - & 94.8 & - & 77.1 & 79.5 & 90.3 \\
 AutoSTR~\cite{zhang2020efficient} & 2020 & 94.7 & 90.9 & - & 94.2 & 81.8 & - & 81.7 & - \\
 Yang \etal~\cite{Holistic} & 2020 & 94.7 & 88.9 & - & 93.2 & 79.5 & 77.1 & 80.9 & 85.4 \\
 SATRN~\cite{SATRN} & 2020 & 92.8 & 91.3 & - & 94.1 & - & 79.0 & 86.5 & 87.8 \\
 SRN~\cite{Yu_2020_CVPR_SRN} & 2020 & 94.8 & 91.5 & 95.5 & - & 82.7 & - & 85.1 & 87.8 \\
 GA-SPIN~\cite{zhang2020spin} & 2021 & 95.2 & 90.9 & - & 94.8 & 82.8 & 79.5 & 83.2 & 87.5 \\
 PREN2D~\cite{PREN} & 2021 & 95.6 & \underline{94.0} & 96.4 & - & 83.0 & - & 87.6 & \underline{91.7} \\
 JVSR~\cite{JVSR} & 2021 & 95.2 & 92.2 & - & \underline{95.5} & - & \underline{\textbf{84.0}} & 85.7 & 89.7 \\
 VisionLAN~\cite{VisionLAN} & 2021 & 95.8 & 91.7 & 95.7 & - & 83.7 & - & 86.0 & 88.5 \\
 ABINet~\cite{ABINet} & 2021 & \underline{96.2} & 93.5 & \underline{97.4} & - & \underline{86.0} & - & \underline{89.3} & 89.2 \\ 
 \midrule
 \multicolumn{2}{l|}{ABINet (reproduced)} & 96.2 & 93.7 & 97.2 & 95.4 & 85.9 & 82.1 & 89.3 & 89.0 \\
 [-0.3em] & & \scriptsize $\pm$0.2 & \scriptsize $\pm$0.4 & \scriptsize $\pm$0.2 & \scriptsize $\pm$0.2 & \scriptsize $\pm$0.2 & \scriptsize $\pm$0.1 &\scriptsize $\pm$0.4 & \scriptsize $\pm$0.3 \\
 \multicolumn{2}{l|}{MATRN (ours)} & \textbf{96.6} & \textbf{95.0} & \textbf{97.9} & \textbf{95.8} & \textbf{86.6} & 82.8 & \textbf{90.6} & \textbf{93.5} \\ 
 [-0.3em]
 & & \scriptsize $\pm$0.1 & \scriptsize $\pm$0.2 & \scriptsize $\pm$0.1 & \scriptsize $\pm$0.1 & \scriptsize $\pm$0.1 & \scriptsize $\pm$0.0 & \scriptsize $\pm$0.2 & \scriptsize $\pm$0.6 \\
 
\bottomrule
\end{tabular}
\end{centering}
\end{table}

Table \ref{tab:accuracy} shows the existing STR methods and their performances on the eight STR benchmark datasets, including the variant versions of IC13 and IC15. 
In this comparison, we only consider the existing methods that are trained on MJSynth and SynthText.

When comparing the existing STR methods, PREN2D, JVSR, and ABINet showed state-of-the-art performances (See underlined values in the table). When compared to them, MATRN achieves the state-of-the-art performances on all evaluation datasets except IC15$_\text{L}$. Specifically, our model achieved superior performance improvements on SVTP and CUTE, 1.3 percent point (pp) and 1.8pp respectively, because these datasets contain low-quality images, curved images, or proper nouns. Therefore, we found that our multi-modal fusion modules resolve the difficulties of scene text images, which cannot solve alone. JVSR~\cite{JVSR} still holds the best position on IC15$_\text{L}$, but MATRN shows huge performance gains on the other datasets: 1.4pp on IIIT, 2.8pp on SVT, 0.3pp on IC13$_\text{L}$, 4.9pp on SVTP, and 3.8pp on CUTE.

For apples-to-apples comparisons, we reproduced ABINet, which is one of the state-of-the-art methods and also our baseline before adding multi-modal fusion modules. In the sanity check, we observed that all reproduced performances are aligned upon confidence intervals from the reported scores. When comparing MATRN from the reproduced ABINet, the performance improvements are statistically significant over all datasets and the gaps are 0.4pp, 1.3pp, 0.7pp, 0.7pp, 1.3pp, and 4.5pp on IIIT, SVT, IC13$_\text{S}$, IC15$_\text{S}$, SVTP, and CUTE, respectively.

Many previous works, such as SE-ASTER, SRN, ABINet, JVSR, and VisionLAN, also analyzed how semantic information can be utilized for text recognition. When compared to them, MATRN shows the best performances on all but one datasets. This result implies that our incorporation method for visual and semantic features is effective compared to the existing methods that utilized semantic information.

\subsection{Performance Comparison under the Comparable Resources}

\begin{table}[t]
\tabcolsep=0.1cm
\caption{Comparison of ABINet and MATRN under the comparable resources. \textbf{\textit{Param.}} indicates the number of model parameters (M) and \textbf{\textit{Time}} represents the inference time (ms/image) with batch size of 1 under AMD 32 cores, RTX 3090 GPU, and SSD 2TB. The underline indicates the similar or more resource when comparing from those of MATRN and the bold represents the best performer.}
\begin{adjustbox}{width=1\linewidth,center}
\label{tab:fair}
\small
\centering
\begin{centering}
\begin{tabular}{l|cc|cccccc|c}
\toprule
 \textbf{Model} & \textbf{\textit{Param.}} & \textbf{\textit{Time}} & IIIT & SVT & IC13$_{\text{S}}$ & IC15$_{\text{S}}$ & SVTP & CUTE & Total. \\
\midrule
 ABINet & 36.7M & 21.6ms & 96.2 & 93.7 & 97.2 & 85.9 & 89.3 & 89.0 & 92.6 \\ \midrule
 ABINet w/ VM-\textit{Big} & \underline{46.2M} & 22.6ms & 96.4 & 93.8 & 97.9 & 86.3 & 89.5 & 88.5 & 92.9  \\
 ABINet w/ LM-\textit{Big} & \underline{46.2M} & 26.6ms & 96.0 & 94.3 & 97.5 & 86.3 & 89.9 & 88.9 & 92.8\\  \midrule
 ABINet w/ VM-\textit{Bigger} & \underline{90.3M} & \underline{29.7ms} & 96.3 & 94.9 & \textbf{98.0} & 86.5 & 89.9 & 89.9 & 93.1 \\ 
 ABINet w/ LM-\textit{Bigger} & \underline{52.5M} & \underline{30.0ms} & 96.1 & 94.3 & 97.7 & 86.0 & 90.1 & 89.2 & 92.8 \\ \midrule
 MATRN (ours) & \underline{44.2M} & \underline{29.6ms} & \textbf{96.6} & \textbf{95.0} & 97.9 & \textbf{86.6} & \textbf{90.6} & \textbf{93.5} & \textbf{93.5} \\
  [-0.3em]
 & & & \scriptsize $\pm$0.1 & \scriptsize $\pm$0.2 & \scriptsize $\pm$0.1 & \scriptsize $\pm$0.1 & \scriptsize $\pm$0.2 & \scriptsize $\pm$0.6 & \scriptsize $\pm$0.1 \\
\bottomrule
\end{tabular}
\end{centering}
\end{adjustbox}
\end{table}

Since MATRN employs additional layers and modules upon ABINet, the performance gains might be considered as the effect of the additional memories and computational costs. To prove the pure benefits of the proposed methods, we evaluate large ABINets that utilize additional memories and computational costs as much as that MATRN requires. Specifically, the scale-up is conducted in two parts; adding transformer layers into VM (or LM) until the models have a similar number of parameters (\textit{Big} models) and a similar inference speed (\textit{Bigger} models). 
Table \ref{tab:fair} shows the evaluation results. The \textit{Big} models have similar parameters with MATRN but their speeds are faster since there is no cross-references between visual and semantic features. By scaling up the models, the \textit{Bigger} models have similar inference speeds with MATRN but hold more parameters. When comparing the performances, the \textit{Big} models provide relatively small performance improvements: 0.3pp of VM-\textit{Big} and 0.2pp of LM-\textit{Big} in Total. The \textit{Bigger} models show better performance improvements than the \textit{Big} models; 0.5pp of VM-\textit{Bigger} and 0.2pp of LM-\textit{Bigger} in Total. However, the performance gains from the scaling-up are restricted when comparing the performance improvements of MATRN; 0.9pp in Total. In addition, the performances of MATRN are statistically significant when comparing all large ABINets. The experiments prove that the benefits of MATRN have not solely lie in the increasing computation resources.

\begin{table}[t]
\caption{Performance improvements through gradual applications of our proposed modules. At the last line, all modules are applied and the model becomes MATRN.}
\begin{adjustbox}{width=1\linewidth,center}
\small
\tabcolsep=0.05cm
\label{tab:abl_study}
\centering
\begin{centering}
\begin{tabular}{r p{0.2\tabcolsep} c p{0.2\tabcolsep} c p{0.2\tabcolsep} c p{0.2\tabcolsep} | cccccc | c }
\cmidrule[\heavyrulewidth]{2-15} 
 && Multi- && SES      && Visual Clue && \multirow{2}{*}{IIIT} & \multirow{2}{*}{SVT} & \multirow{2}{*}{IC13$_{\text{S}}$} & \multirow{2}{*}{IC15$_{\text{S}}$} & \multirow{2}{*}{SVTP} & \multirow{2}{*}{CUTE} & \multirow{2}{*}{Total.} \\
&& modal FE          && Encoding && Masking && &&&&&&\\
\cmidrule{2-15} 
{\scriptsize ABINet$\rightarrow$} &&        &&   &&       && 96.2 & 93.7 & 97.2 & 85.9 & 89.3 & 89.0 & 92.6 \\
\cmidrule{2-15} 
&&\cmark  &&        &&       && 96.5 & 94.3 & 98.0 & 85.9 & 90.1 & 91.0 & 93.0 \\
\cmidrule{2-15} 
&&\cmark  && \cmark &&       && 96.4 & 94.7 & \textbf{98.1} & \textbf{86.9} & 90.4 & 89.9 & 93.3\\
\cmidrule{2-15} 
{\scriptsize MATRN$\rightarrow$} && \cmark  && \cmark &&\cmark && \textbf{96.6} & \textbf{95.0} & 97.9 & 86.6 & \textbf{90.6} & \textbf{93.5} & \textbf{93.5}\\
\cmidrule[\heavyrulewidth]{2-15} 
\end{tabular}
\end{centering}
\end{adjustbox}
\end{table}

\subsection{Ablation Studies on Proposed Modules}
Here, we analyze how the proposed modules contribute to the final performances. Table~\ref{tab:abl_study} shows the ablation studies that start from ABINet and add the proposed modules one by one. As can be seen, the total performances increase when adding proposed modules gradually. The application of the multi-modal transformer provides 0.4pp performance improvements based on ABINet. By applying SES on the multi-modal transformer, the total performance increases by 0.3pp. When adding the visual clue masking, the total performance finally becomes 93.5\% with the 0.2pp improvement. Consequently, the simple application of a multi-modal transformer brings 0.4pp and our novel modules provide 0.5pp of performance improvements. We should note that applying a multi-modal transformer requires additional computations and parameters but the other proposed modules use quite small computations without any additional parameters. The ablation study indicates that our proposed modules for better multi-modal fusion lead to better STR performances effectively.

\subsection{Discussion}

\begin{table}[t]
\small
\caption{Comparisons between uni-modal and multi-model FEs.}
\tabcolsep=0.12cm
\label{tab:multi-modal-fe}
\centering
\begin{centering}
\begin{tabular}{cc|cccccc|c}
\toprule
Visual FE & Semantic FE & IIIT & SVT & IC13$_\text{S}$ & IC15$_\text{S}$ & SVTP & CUTE & Total. \\
\midrule
&  & \textbf{96.5} & 93.2 & 97.0 & 85.9 & 89.0 & 89.2 & 92.7 \\ \midrule
 & \cmark & \textbf{96.5} & 93.8 & 97.2 & 85.8 & 89.0 & 90.3 & 92.8 \\ \midrule 
\cmark & & 96.1 & 93.5 & 97.5 & 86.1 & 89.8 & \textbf{91.3} & 92.8 \\  \midrule 
\cmark& \cmark & 96.4 & \textbf{94.7} & \textbf{98.1} & \textbf{86.9} & \textbf{90.4} & 89.9 & \textbf{93.3} \\ 
\bottomrule
\end{tabular}
\end{centering}
\end{table}

\subsubsection{Uni-modal vs. Multi-modal Feature Enhancement.} The existing methods~\cite{JVSR,VisionLAN} focus on uni-modal FE by utilizing the other modality. To analyze the benefit of multi-modal FE, we compare the uni-modal FE that updates only a single modality utilizing the multi-modal transformer. In this experiment, we use SES for better fusion through the multi-modal transformer but do not apply the visual clue masking strategy for a fair comparison. Table~\ref{tab:multi-modal-fe} provides the comparison results. In the table, the first model is identical to ABINet with SES and the uni-modal FE models (the second and third models) update the target features through the multi-modal transformer. As can be seen, the uni-modal FEs provide marginal performance improvements; 0.1pp in Total. When enhancing both modalities (the last model), the STR model enjoys two benefits of the semantic and visual FE and shows large performance improvements in Total. Given these points, we found that combining visual and semantic features improves the recognition performance, but one-way information flows are not enough to fuse two modalities. Besides, the multi-modal FE enables the two features to communicate in both directions and provides better performance.

\begin{table}[t]
\caption{STR Performances with each level of features from VM, LM, and their fusions. Each value indicates total STR accuracy (\%). $\mathbf{V}$ and $\mathbf{S}$ represent the output features from VM and LM, respectively. $\mathbf{V}^\text{M}$ and $\mathbf{S}^\text{M}$ indicate the enhanced features through the cross-reference. $\mathbf{F}$ represents the combined features for the final output.}
\small
\tabcolsep=0.2cm
\label{tab:final_prediction}
\centering
\begin{centering}
\begin{tabular}{l|cc|cc|c}
\toprule
Model & $\mathbf{V}$ & $\mathbf{S}$ & $\mathbf{V}^\text{M}$ & $\mathbf{S}^\text{M}$ & $\mathbf{F}$ \\
\midrule
ABINet & 90.9 & 49.5 & - & - & 92.6 \\
MATRN & 91.2 & 52.6 & 93.4 & 93.4 & 93.5 \\
\bottomrule
\end{tabular}
\end{centering}
\end{table}

\subsubsection{STR Performance at Each Level of Features}
MATRN utilizes multi-task cross-entropy objectives, described in \S3.7. Here, we evaluate the STR performances from the multiple features; $\mathbf{V}$, $\mathbf{S}$, $\mathbf{V}^\text{M}$, $\mathbf{S}^\text{M}$, and $\mathbf{F}$. Table~\ref{tab:final_prediction} shows the results of ABINet and MATRN. Interestingly, the results of $\mathbf{S}$ show insufficient performances in both models by refining character sequences without consideration of visual clues. However, the semantic features are combined and lead to better performances; $\mathbf{F}$ (ABINet), $\mathbf{V}^\text{M}$, and $\mathbf{S}^\text{M}$ (MATRN). In addition, the multi-modal features in MATRN show better performances than the final performances of ABINet and their combination shows the best. 

\begin{figure}[t]
\centering
    \subfloat{
        \includegraphics[width=0.25\linewidth]{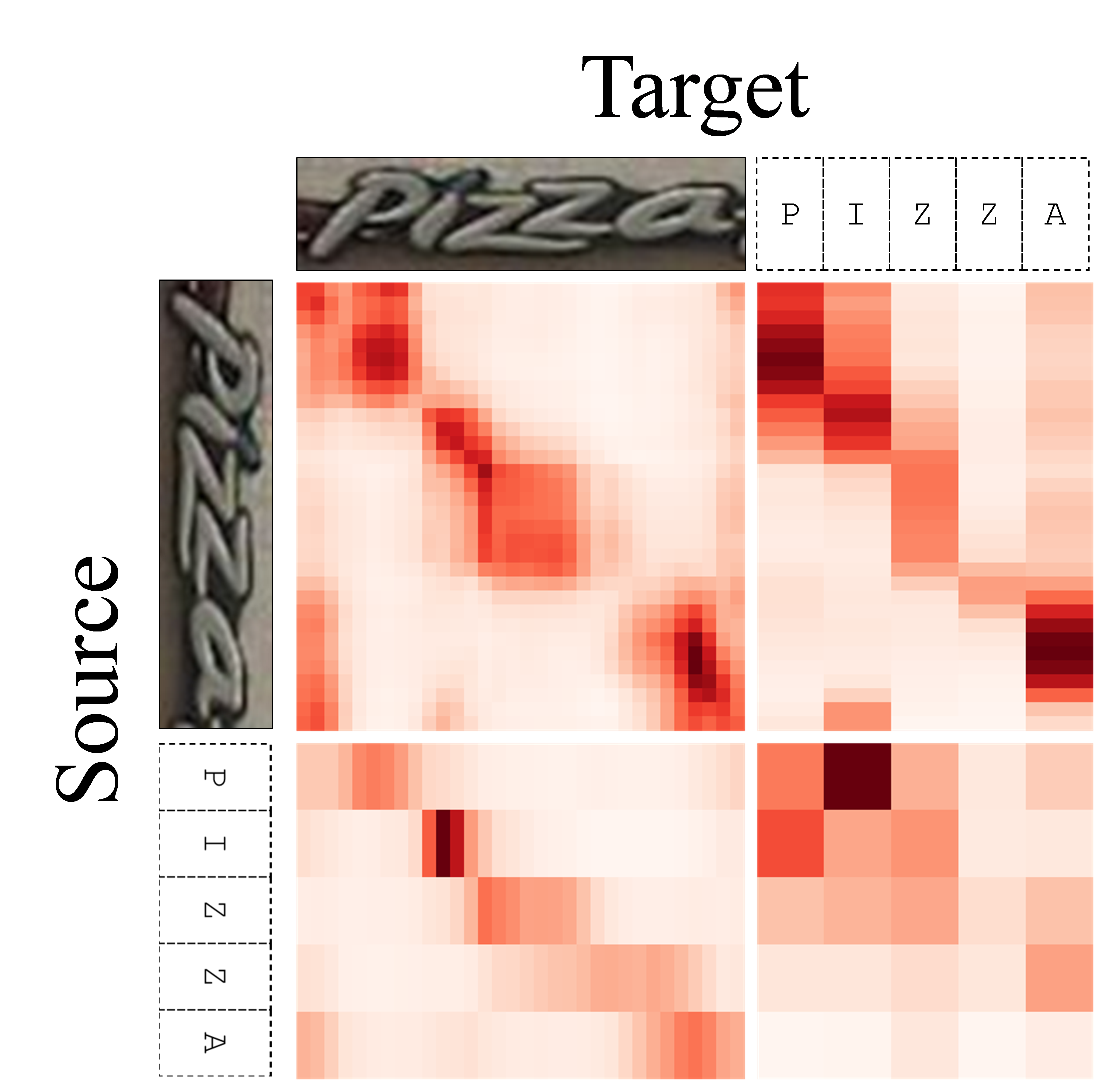}
    }
    \subfloat{
        \includegraphics[width=0.25\linewidth]{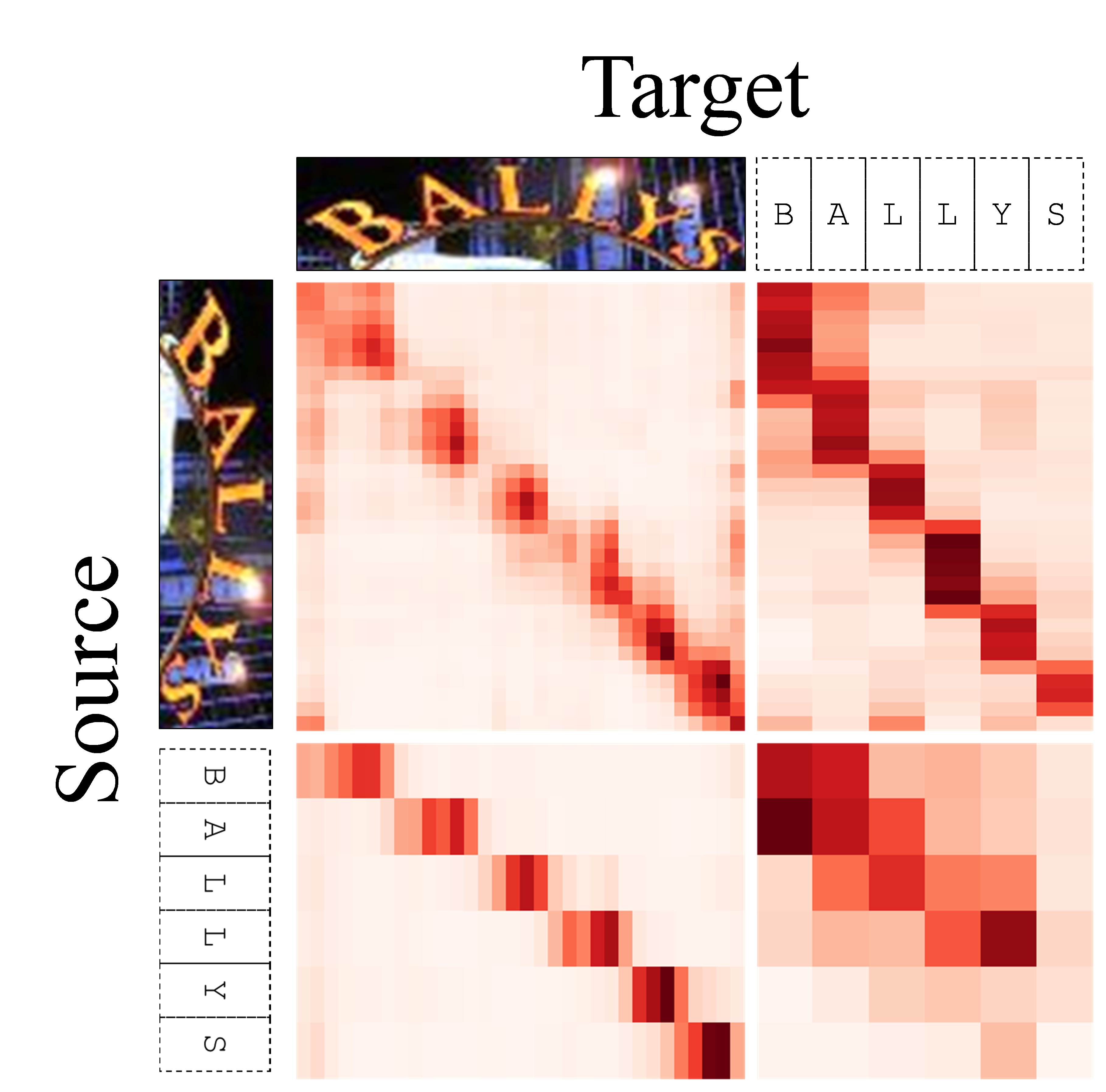}
    }
    \subfloat{
        \includegraphics[width=0.25\linewidth]{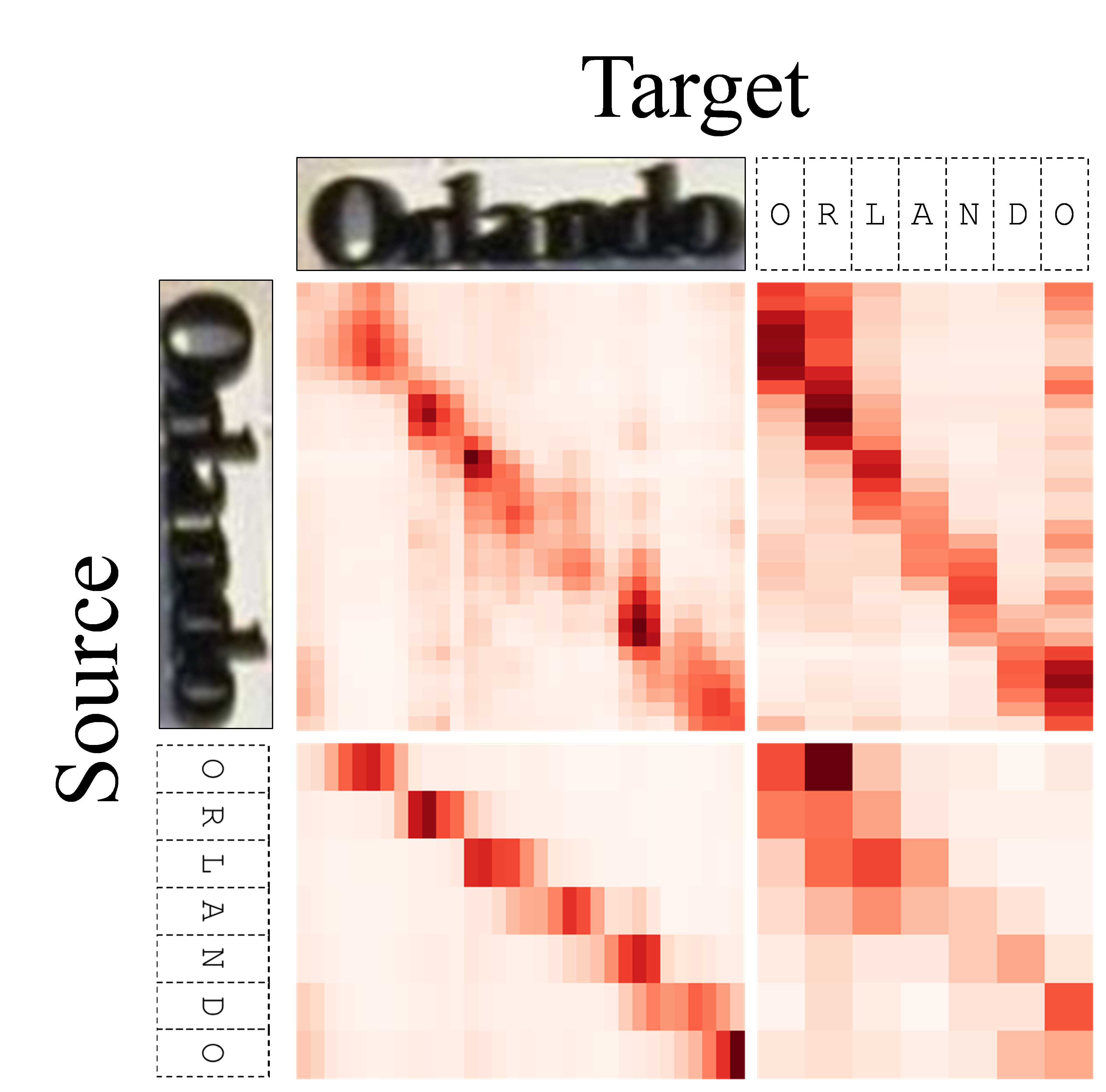}
    }
    \caption{Examples of self-attention maps in multi-modal FE. Attention maps on the top-right and the bottom-left indicate the cross attentions between two modalities.}
    \label{fig:attn_map}
\end{figure}

\begin{figure}[t]
\centering
    \includegraphics[width=0.99\linewidth]{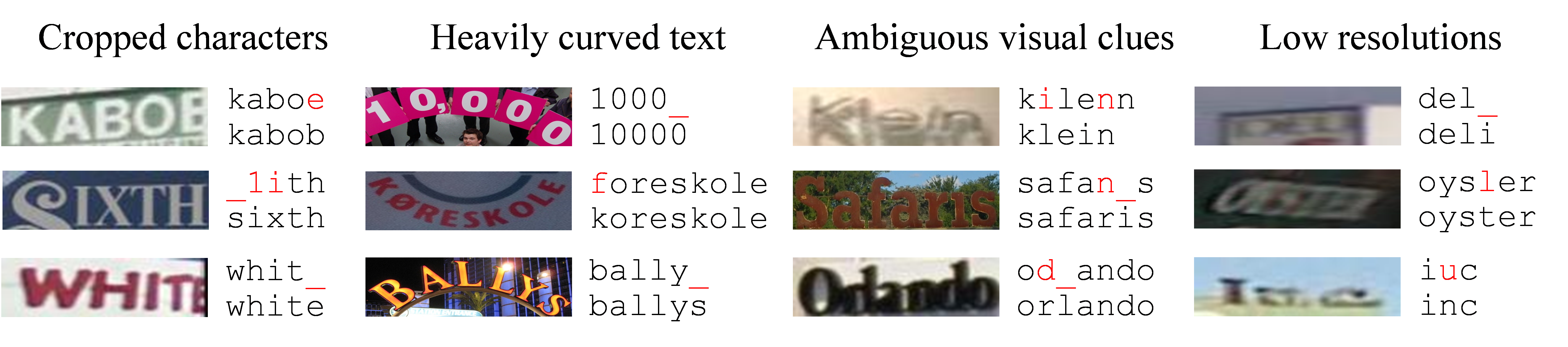}
    \caption{Examples that ABINet fails (\textit{first line}) but MATRN succeeds (\textit{second line}).}
    \label{fig:examples}
\end{figure}

\subsubsection{Analysis on Cross-references}

Figure~\ref{fig:attn_map} shows attention map examples identified by the multi-modal FE of MATRN. At each attention map, the top-left and the bottom-right show the uni-modal attentions referring to their uni-modal features and the others provide the cross attentions between two different modalities. As can be seen in the examples, visual and semantic features refer to their own modality as well as interact with each other. 

\subsubsection{Analysis on Previous Failure Cases}

Figure~\ref{fig:examples} shows the test examples that ABINet fails but MATRN successes. As can be seen, MATRN provides robust results on ``cropped characters'', ``heavily curved text'', ``ambiguous visual clues'' and ``low resolutions''. The results show that MATRN tackles the existing challenges.

\section{Conclusion}

This paper explores the combinations of visual and semantic features identified by VM and LM for better STR performances. 
Specifically, we propose MATRN that enhances visual and semantic features through cross-references between two modalities.
MATRN consists of SES, that matches semantic features on 2D space that visual features are aligned in, multi-modal FE, that updates visual and semantic features together through the multi-modal transformer, and visual clue masking strategy, that stimulates the semantic references of visual features. In our experiments, naive applications of the multi-modal transformer lead to marginal improvements from the baseline. To this end, the components of MATRN effectively contribute to the multi-modal combinations and MATRN finally achieves state-of-the-art performances on seven STR benchmarks with large margins.

\clearpage
%
%
\bibliographystyle{splncs04}
\bibliography{main}
\end{document}